\documentclass[conference]{IEEEtran}
\IEEEoverridecommandlockouts
\usepackage{cite}
\usepackage{amsmath,amssymb,amsfonts}
\usepackage{algorithmic}
\usepackage{graphicx}
\usepackage{textcomp}
\usepackage{xcolor}

\usepackage{hyperref}
\usepackage{multirow}
\usepackage{graphicx}
\usepackage{amsmath}
\usepackage{amsmath, balance}
\usepackage{scalerel}
\usepackage{tikz}
\usetikzlibrary{svg.path}
\usepackage{arydshln}
\usepackage{caption}
\usepackage{academicons}
\usepackage{tabularray}  
\usepackage{lipsum}  
\usepackage{adjustbox}
\usepackage{mdframed}
\usepackage{orcidlink}
\usepackage{svg}

\definecolor{Gr}{rgb}{0.0, 0.5, 0.0}
\definecolor{light-gray}{gray}{0.95}
\newcommand{\code}[1]{\colorbox{light-gray}{\texttt{#1}}}

\def\BibTeX{{\rm B\kern-.05em{\sc i\kern-.025em b}\kern-.08em
    T\kern-.1667em\lower.7ex\hbox{E}\kern-.125emX}}

\begin{document}

\title{StackRAG Agent: Improving Developer Answers with Retrieval-Augmented Generation}

\author{\IEEEauthorblockN{Davit Abrahamyan}
\IEEEauthorblockA{\textit{Department of Computer Science} \\
\textit{University of British Columbia}\\
Kelowna, Canada \\
0009-0004-4397-1699}
\and
\IEEEauthorblockN{Fatemeh H. Fard}
\IEEEauthorblockA{\textit{Department of Computer Science} \\
\textit{University of British Columbia}\\
Kelowna, Canada \\
fatemeh.fard@ubc.ca}
}

\maketitle

\begin{abstract}
Developers spend much time finding information that is relevant to their questions. Stack Overflow has been the leading resource, and with the advent of Large Language Models (LLMs), generative models such as ChatGPT are used frequently. However, there is a catch in using each one separately. Searching for answers is time-consuming and tedious, as shown by the many tools developed by researchers to address this issue. On the other, using LLMs is not reliable, as they might produce irrelevant or unreliable answers (i.e., hallucination). In this work, we present StackRAG, a retrieval-augmented Multiagent generation tool based on LLMs that combines the two worlds: aggregating the knowledge from SO to enhance the reliability of the generated answers. Initial evaluations show that the generated answers are correct, accurate, relevant, and useful. A description video can be found \href{https://bit.ly/3xxmWAm}{here}\footnote{Please note that as the tool requires API keys, we are not able to set up a live demo of StackRAG.}.
\end{abstract}


\section{Introduction}
In the recent year of technological innovations, the groundbreaking improvements in Natural Language Processing, epitomized by Large Language Models (LLM) such as GPT and Llama have revolutionized the way people interact with technology. Developers increasingly rely on such tools to address various challenges faced during the software development process, including but not limited to code generation and bug detection~\cite{bugs}. 
However, such models are limited by their static training data. As a result, they cannot keep up with the recent innovations and, thus, cannot provide up-to-date responses to the latest challenges~\cite{outdated}. This is specifically more challenging for the field of software engineering, which has many changes as new APIs, libraries, or frameworks are introduced. 
Moreover, the possibility of hallucinations poses a significant problem of their reliability in the software development process~\cite{survey}, where bugs can result in fatal consequences~\cite{bugrisks}.

On the other hand, using software engineering techniques is recommended to develop reliable LLM-based models/tools~\cite{survey}. 
Thus, even though the emergence of LLMs has ultimately improved the software development process, the usage of widely known platforms such as Stack Overflow (SO) is still prevalent and indispensable. SO provides a medium for developers to access solutions to challenges faced by other developers, engage in meaningful discussions, and facilitate the exchange of expertise between developers~\cite{sobenefits}. By utilizing the collective knowledge available in SO, developers can stay up-to-date with the challenges that the latest advancements pose, relying on the expertise and knowledge of other developers. 
However, searching for related posts to explore solutions is tedious and time-consuming~\cite{howFar} and only relying on LLMs is not reliable~\cite{survey}. 

Thus, in this paper, we propose a Retrieval Augmented Generation (RAG)-based Multiagent LLM tool called \textbf{StackRAG}.
StackRAG utilizes the public knowledge of the developers' community from SO and combines it with the linguistic abilities of GPT to provide a tool that answers developers' queries reliablity and with up to date information; addressing the current challenges of LLM usage. 
Our agent-LLM tool aims to provide developers with more grounded and accurate answers, which will result in increased efficiency of the software development process.

Though agent-based models have been developed for various other tasks (see Section~\ref{sec:relatedWorks}), there is no study or tool that is similar to our work in the context of providing accurate and useful knowledge from SO, with LLM-based models. 
Our initial evaluations show that compared to the base LLM, GPT 4, StackRAG provides more correct, accurate, relevant, and useful responses. 
We open-source the tool\footnote{\url{https://github.com/DavidAbrahamyan/StackRAG}}.



\section{Related Works} \label{sec:relatedWorks}

Mining Stack Overflow has a long history in software engineering research. SO has been studied for the information highlights where the authors developed a recommender system for content highlighting with formatting styles~\cite{ahmed2024studying}, API recommendation~\cite{wei2022clear}, representation learning for various tasks~\cite{howFar}, tag recommendation~\cite{he2022ptm4tag}, recommending code snippets~\cite{recommendations}, and finding question relatedness~\cite{pei2021attention}. 
Our work aims to find related SO posts for a given query. In this sense, our study is more similar to the `relatedness' studies. 
However, StackRAG differs from all the relatedness and recommendation systems that are developed for SO in several aspects. 
We use the Multiagent LLM-based paradigm, which makes the user's process from searching to response generation seamless. 
Our approach combines the knowledge source from Stack Overflow with the generation capability of language models. This innovative direction of incorporating agents in software engineering is a significant step forward in the field. 

The other related category of studies is the one that uses multiagent or agent-based paradigms. Codeagnet~\cite{tang2024codeagent}, RepoAgent~\cite{luo2024repoagent}, and RepairAgent~\cite{bouzenia2024repairagent} are examples of such research in the software engineering domain. 
PaperQA~\cite{paperqa} is a similar work, a RAG-based agent capable of answering questions related to scientific papers. 
These studies differ from our work not only in the application area and context used but also in the processes used to extract relevant posts. 



\begin{figure*}
    \centering
    \includegraphics[width=\textwidth]{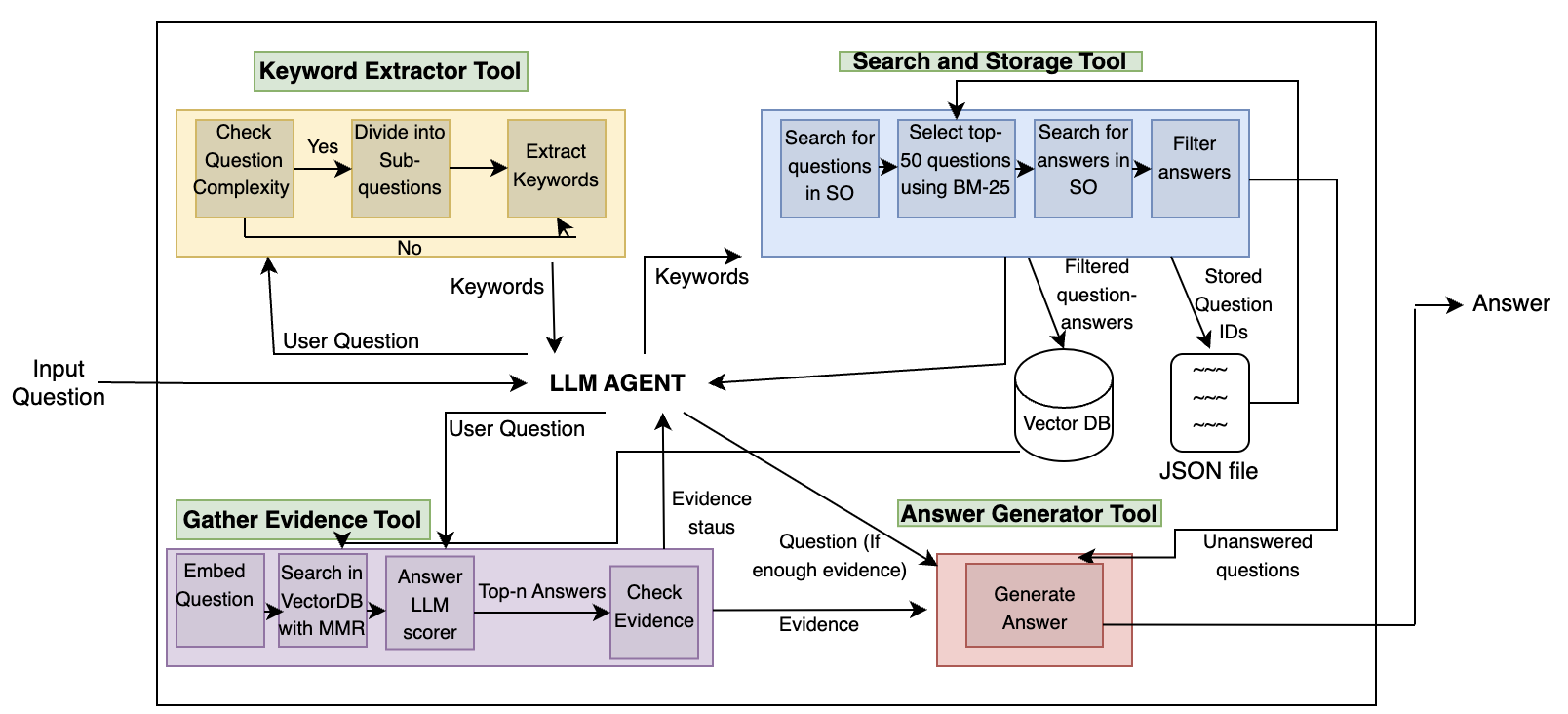}
    \caption{Architecture Diagram of StackRAG and LLM-based agent designed to foster the software development process.}
    \label{fig:architecture}
\end{figure*}

\section{Architecture}

StackRAG is a Multiagent LLM-based tool developed using the LangChain Agent framework~\footnote{https://js.langchain.com/v0.1/docs/modules/agents/}. StackRAG is designed to help developers find solutions to the problems they face by utilizing the capabilities of GPT and the knowledge available on Stack Overflow. 
The overall architecture of StackRAG is provided in Figure~\ref{fig:architecture}. 
StackRAG is equipped with four components, each specialized in providing a specific type of functionality. 
The components are the Keyword Extractor, Search and Storage component, Evidence Gatherer, and Answer Generator. 
Except for the Search and Storage component, the other ones include agents handling their objectives and other processing.
A user can ask a question $Q$ from StackRAG and the central orchestrator agent is responsible for deciding which component to use. 
StackRAG's evidence-gathering process is comprehensive and meticulous. It uses keywords extracted from the question to locate relevant question-answer pairs from Stack Overflow. After a series of filtering and processing steps, the most pertinent question-answer pairs are gathered as evidence. If the collected evidence is deemed sufficient, the agent proceeds to answer the question. If not, the process restarts until the necessary amount of evidence is gathered, ensuring a thorough and accurate response.
GPT is the base language model used as an agent in all components. 
We explain the details of each component below.

\textbf{Keyword Extractor}. 
This component is responsible to extract keywords based on user's query. The extracted keywords identify the main concepts of the query and will be used in later stages to find related questions and answers from Stack Overflow. 
The query is processed initially by question\_complexity\_checker agent to assess if it should be broken down into sub-questions. 
The reason is that if the query is long or complicated, we might miss some important concepts if we extract keywords directly. 
Therefore, we assess the query first by asking the agent, and based on the boolean value of its complexity evaluator, the output is a list of sub-questions or is sent to the keyword\_extractor agent. 
In the case of sub-questions, they are passed into keyword\_extractor asynchronously to speed up the process. 
The question\_complexity\_checker and keyword\_extractor agents are using the following prompts to accomplish their tasks. 
The list of keywords $K = {k_1, k_2, ... , k_n}$ from this component is sent to the central agent for further processing.

\noindent \code{PROMPT (question\_complexity\_checker):} \\ 
\textit{You are a part of RAG architecture that specializes in generating answers to user's given query using Stack Overflow.
You are going to be provided the user question. Your task is to determine whether the question is complex enough to be divided into sub-questions.
If, in order to answer the question, different topics have to be covered, return TRUE, all in capital letters. If there are multiple simple questions, in the given question, again, return TRUE. Otherwise, if you think that the question is not complex and there is no need to divide it into sub-questions, return FALSE.
Do not provide explanations for your choice, t=output a single word, either TRUE or FALSE.
Question: {question}}

\noindent \code{PROMPT (keyword\_extractor): } \\ 
\textit{You are a question-to-query parser. You are given a technical question. You have to use the question to create a Python list of search queries that will be useful in conducting a search in Stack Overflow. Make every query in the list as short as possible. Having less words will produce better results. But make sure you do not omit important search terms and make the search query too general. It does not have to be a complete sentence. Every single query in the list MUST be less than 4 words. Output MUST be a Python list with every element enclosed with double quotes. Question: {question}}


\textbf{Search and Storage}. 
This component is responsible for searching and retrieving questions and answers from Stack Overflow related to the user's query.
We search for relevant questions and their answers separately, as this approach could diversify the related SO posts, and more relevant information could be extracted. 
For this purpose, the list of extracted keywords $K$ are provided and a search is conducted in Stack Overflow, using the StackExchange API. 
Initially, we designed our system to search for questions and corresponding answers asynchronously, which would have reduced the time spent gathering the answers. However, because StackExchange API heavily penalizes users who make more than 30 API calls per second, we decided to retrieve the results sequentially.


Given a list of keywords $K$, we store the information of the retrieved questions, including the question's \texttt{ID}, \texttt{LINK}, \texttt{TITLE}, \texttt{BODY}, \texttt{CREATION DATE}, and \texttt{ACCEPTED-ANSWER-ID} or \texttt{NONE} if there is no accepted answer. 
In order to find the relevant questions to the user's query and eliminate irrelevant questions, we use \textit{BM-25} re-ranking algorithm~\cite{bm25}. The Stack Overflow questions' \texttt{TITLE} and \texttt{BODY} are fed to BM-25 as a single string, and the user's question is given as the query. We use the top 50 questions from the output of BM-25. The top 50 are chosen based on our empirical experiments. 
This step leaves us with the top 50 most relevant questions, which are divided into two lists: the ones that contain an accepted answer and the ones that do not. 
The questions with accepted answers are used in the next other components to provide evidence for answer generation. 
The questions with no accepted answer are stored for the final answer generation, which are provided as links in the generated answer. The rationale is that these questions or some of their replies could potentially be useful for users, or they might be interested in investigating those SO posts. 



The question \texttt{ID}'s of the ones having an accepted answer are then used to search for their corresponding answers on SO. 
A question can have an accepted answer as well as many unaccepted answers, which are still relevant to the question and could provide some insight or useful information for the user's query. Therefore, we collect the accepted answers and the top 2 unaccepted answers to store in our vector database. 
For collecting the unaccepted answers, we use two criteria: the $score$ that each answer has in SO, and the $creation date$ of the answer. 
We choose answers with higher $score$s and the ones that are more recent.


We store the extracted questions and answers as vector embeddings using text-embedding-ada-3-small model developed by OpenAI\footnote{https://platform.openai.com/docs/models/embeddings}
The questions' \texttt{TILE} and \texttt{BODY} are combined with their retrieved answers (i.e., accepted and top-2 unaccepted answers) to generate its embedding for storage in Pinecone Vector Database. 
Vector embeddings are the vector representations of data in n-dimensional space. In order to avoid the necessity of constantly calculating the embedding vectors of the same answers that we retrieve from SO, we are using Pinecone Vector Database\footnote{https://www.pinecone.io/}. In vector databases, each piece of data is stored as a vector rather than as is. Such databases often offer searching functionalities to users, such as cosine similarity when searching for input.


Additionally, we store the \texttt{ID}'s of the questions in Pinecone in a JSON file locally. This local JSON file is intended to eliminate potential redundant calls to StackExchange API if a similar question is asked by the user. 
Therefore, with each new query, before searching in SO, we check in the JSON file to see if we have already retrieved the answer previously. 
This process reduces the number of calls and, therefore, the tool's overall query-response time.
After all, a message is returned to our agent indicating if the data has been successfully stored in the database. If successfully stored, the agent proceeds to the next step. Otherwise, it repeats the searching and storing process. If searching fails after multiple trials, the tool generates an output noting that no results are found. 



\textbf{Evidence Gatherer}. 
The Evidence Gatherer component uses the data stored in the Pinecone database to gather the most relevant pieces of information, which is used to answer the user's query. 
For this purpose, first we compute the similar question-answer pairs that also include diverse set of information. Then, we calculate their relevancy to the user's query. 
The results of these two steps gives us $n$ question-answer pairs that is given to the agent as evidence. 
The agent should identify whether such evidence is sufficient to answer the query or the whole process (including keyword extraction) should be repeated. 
We use $n=3$ in our work, which is set empirically.
The details of these steps are explained below. 

To find similar question-answer pairs, first, the embedding of the user's query is generated using the text-embedding-ada-002 model. 
This embedding is compared with the embedding of question-answer pairs stored in the Pinecone database using the cosine similarity metric.
We do not use a threshold for the similarity score as we are interested to find the top results not necessarily the most similar one above a specific percentage.  
Based on the cosine similarity score, we save the top 30 results and re-rank them using Maximum Marginal Relevance (MMR)~\cite{mmr}.
As there can be multiple questions on SO containing similar questions and answers, we use MMR to select SO posts that are most similar and optimize for diversity in the retrieved answers. This diversity in the extracted question-answer pairs will help the model generate more well-grounded answers.
After applying MMR, we keep the top 15 results that will be used as evidence. 

After gathering the evidence, we use evidence\_scorer agent to score each piece of evidence (i.e., each question-answer pair) based on their relevance to the user's query. The evidence\_scorer agent provides a score ranging from $1$ to $5$, assessing how relevant the evidence is in order to answer the question. A score of $5$ indicates high relevance, and $1$ indicates low relevance. If the retrieved evidence is not relevant at all, the agent returns ``not useful'' for that piece of evidence. To speed the process, the scoring of each piece of evidence is done in parallel, in asynchronously with other tasks. 
Finally, we choose the top 3 pieces of evidence and we provide it as a single full evidence, $E$, which is provided to the evidence\_checker agent. 
The evidence\_checker agent returns a boolean value indicating if the gathered evidence $E$ is sufficient in order to answer query $Q$.
If the gathered Evidence $E$ is enough, we proceed to generating the answer using Answer Generator agent. Otherwise, the tool should repeat the process from the beginning and proceeds to generate a new set of keywords.
The prompts used for evidence\_scorer and evidence\_checker agents are as follows. 

\noindent \code{PROMPT (evidence\_scorer): } \\ 
\textit{
You are a part of RAG architecture that specializes in generating answers to user's given query using Stack Overflow.
Provided the gathered evidence from Stack Overflow as well as the user's given question, your task is to determine how useful the evidence is in order to answer the user question. The evidence includes a question and its corresponding answer from Stack Overflow. Rate the given evidence on the scale from 1 to 5, with 1 indicating not useful and 5 indicating really useful. If the evidence is not useful at all, return ``not useful" all in lowercase. Only output either a number from 1-5 or ``not useful" with no explanation.
Gathered Evidence:
\{evidence\}
User Question:
\{question\}
}

\noindent \code{PROMPT (evidence\_checker): } \\ 
\textit{You are a part of RAG architecture that specializes in generating answers to user's given query using Stack Overflow.
Provided the gathered evidence from Stack Overflow as well as the user's given question, your task is to determine whether you have enough evidence to answer the question or not.
Do not generate answer even if you have enough evidence. The evidence does not have to directly answer the question, but it has to provide the basis upon which you can form the answer. If no such evidence is provided, return ``FALSE", do not use your own knowledge to answer the question.
Your output must be a single word, either ``TRUE" or ``FALSE". All letters must be capital, do not explain why you chose a specific answer, only output either ``TRUE" or ``FALSE"
Gathered Evidence: \{evidence\}
User Question: \{question\}}

\textbf{Answer Generator}. 
The main responsibility of this component, which is composed of a single agent, is to generate an answer given the evidence $E$ and user's query $Q$. The agent is instructed to always provide the links to the questions that were used in the response, so that the user can access and use all the responses available on SO. 
Additionally, we provided it with the links to the questions that did not have an accepted answer but were relevant to the question (Stored in the Search and Storage Tool after applying BM-25). 
Empirically, we found that providing links to such questions is a helpful feature for the user, as they can track those links and find useful answers in the future.
The prompt for the answer\_generator agent is: 

\noindent \code{PROMPT (answer\_generator): } \\ 
\textit{
You are a part of RAG architecture that specializes in generating answers to user's given query using Stack Overflow.
You are the final piece of this architecture, your task is to construct the final answer based on the given question and the provided evidence.
Be as thorough as possible, if you write code, do not omit anything, write every single detail.
Indicate whether the answer that you used in generating the response was an accepted answer in Stack Overflow or not.
At the end of your answer, mention all the links of the answers that you used in the following format:\\
Links used:\\
- [Question Title] Link1\\
- [Question Title] Link2\\
- [Question Title] Link3\\
...\\
You will also be provided a list of questions which are unanswered but are relevant to the user query, include their links at the end in the following format:\\
Unanswered questions that you may find useful in the future:\\
- [Question Title] Link1\\
- [Question Title] Link2\\
- [Question Title] Link3\\
...\\
User Question: \{question\}
Gathered Evidence: \{evidence\}
Unanswered Question List: \{unanswered\_question\_list\}
}

\section{Evaluation}

To evaluate StackRAG's performance, we conducted a manual evaluation by three software developers with different titles: Machine Learning Engineers, NLP engineers, and full-stack developers. 
These developers have more than five years of experience and often use ChatGPT to find answers to their development-related questions. 
We selected three questions that developers have struggled with in their careers and generated answers using GPT-3.5, GPT-4, and StackRAG. 
We asked developers to evaluate the generated answers based on the metrics discussed below.  
The three challenged questions are 
\textit{`How to do horizontal scaling of web sockets?'}, 
\textit{`How to import from a parent directory in Python?'}, and
\textit{`How to use function calling in openAI API?'}.

\textbf{Evaluation Metrics.}
We used four different evaluation metrics to assess the performance of our model. Each metric has a score between $1$ to $5$, one being the worst and 5 indicating the best. 

\textit{Correctness (C)}: Is the response provided to the question correct?

\textit{Accuracy (A)}: Does the response accurately explain the question's solution?

\textit{Relevance (R)}: Is the generated text related to the asked question in terms of the topic?

\textit{Usefulness (U)}: Does the response provide useful information to solve the question?




\textbf{Results.}
Table~\ref{tab:comparison-gpt} shows the evaluation of StackRAG, GPT-3.5, GPT-4 models, reporting the average scores the developers gave to the generated answers of the models.
Even though StackRAG uses GPT-4 as the base model, it performs better than GPT-4 in all four metrics.  
This indicates that the developers find the answers generated by StackRAG more relevant to the topic. In addition to correct and accurate answers, they found more useful information in the StackRAG's responses. 
When we discussed this with the developers, they valued the links in the generated response.

\begin{table}[h!]
  \centering
  \begin{tabular}{|c|c|c|c|c|}
    \hline
    \textbf{Metric/Model} & \textbf{C} & \textbf{A}  & \textbf{R}  & \textbf{U}\\
    \hline
    \textbf{StackRAG} & 4.66 &  4.55 & 4.55 & 4.55 \\
    \hline
    \textbf{GPT-3.5} & 4.22 &  4.11 & 4 & 3.77 \\
    \hline
    \textbf{GPT-4} & 4.22 &  4.11 & 4.22 & 4.11 \\
    \hline
  \end{tabular}
  \caption{Evaluation of StackRAG and GPT models.}
  \label{tab:comparison-gpt}
\end{table}

\section{Limitations and Improvements}

One of StackRAGl's main limitations is that it takes more time than directly prompting GPT models, as it uses multiple agents and a few calls to the base LLM. 
Another reason for the increased response time is the calls to SO API, which has a daily limit of 10000 API calls. We have tried to mitigate the time issue by adding more asynchronous processes to search for answers. However, SO heavily penalizes us for making multiple API calls in a short period, which has reduced our total number of calls per day even more. 
We are working on techniques to replace some agents or reduce the LLM calls to decrease the response time.  
However, note that we should ``get used to delegating tasks to AI agents and potentially wait for a response''~\footnote{\url{https://www.youtube.com/watch?v=sal78ACtGTc&t=591s&ab_channel=SequoiaCapital}}. So we should be prepared to wait for longer times, as the overall workflow is made easier and reduces significant efforts from the developers. When StackRAG is used, instead of searching on a search engine and investigating through several links and answers, or instead of several iterations of the prompts with an LLM and validating the responses, StackRAG provides a comprehensive, reliable response with links at once.  


\section{Conclusion}

The new paradigm in software engineering is considered the usage of LLM-based agents. In this work, we present StackRAG, a Multiagent RAG-based tool that enhances the developers' experience when searching for a query from Stack Overflow. StackRAG aims to generate reliable answers based on SO, instead of directly asking an LLM. The initial evaluations show the potential benefit of StackRAG when searching for related queries from SO, compared to GPT-family models. A future direction is to enhance the tool by adding search engines and other repositories to provide the answers. 


\balance

\bibliographystyle{IEEEtran}
\bibliography{bibliography}

\end{document}